\documentclass{article}

\usepackage{PRIMEarxiv}

\usepackage[utf8]{inputenc} 
\usepackage[T1]{fontenc}    
\usepackage{hyperref}       
\usepackage{url}            
\usepackage{booktabs}       
\usepackage{amsfonts}       
\usepackage{amsmath}
\usepackage{nicefrac}       
\usepackage{microtype}      
\usepackage{lipsum}
\usepackage{fancyhdr}       
\usepackage{graphicx}
\usepackage{multirow}
\graphicspath{{media/}}     

\title{Lifting Vision: Ground to Aerial Localization with Reasoning Guided Planning}

\author{%
Soham Pahari\textsuperscript{1} \quad
M Srinivas\textsuperscript{2} \\[1.0ex]
\textsuperscript{1}School of Computer Science, UPES \quad
\textsuperscript{2}Department of CS\&E, NIT Warangal
\\ [1.0ex]
\texttt{soham.109424@stu.upes.ac.in} \quad
\texttt{msv@nitw.ac.in}
}

\begin{document}
\maketitle

\begin{abstract}
Multimodal intelligence development recently show strong progress in visual understanding and high level reasoning. Though, most reasoning system still reply on textual information as the main medium for inference. This limit their effectiveness in spatial tasks such as visual navigation and geo-localization. This work discuss about the potential scope of this field and eventually propose an idea visual reasoning paradigm Geo-Consistent Visual Planning, our introduced framework called Visual Reasoning for Localization, or ViReLoc, which performs planning and localization using only visual representations. The proposed framework learns spatial dependencies and geometric relations that text based reasoning often suffer to understand. By encoding step by step inference in the visual domain and optimizing with reinforcement based objectives, ViReLoc plans routes between two given ground images. The system also integrates contrastive learning and adaptive feature interaction to align cross view perspectives and reduce viewpoint differences. Experiments across diverse navigation and localization scenarios show consistent improvements in spatial reasoning accuracy and cross view retrieval performance. These results establish visual reasoning as a strong complementary approach for navigation and localization, and show that such tasks can be performed without real time global positioning system  data, leading to more secure navigation solutions.

\end{abstract}


\section{Introduction}
In this current era of large language models(LLM) \cite{rw2_2, rw2_7, rw2_41} and at a extend to that multimodal large language models(MLLM)\cite{rw2_46, rw2_25}, visual understanding has improved at a solid pace. These models show strong perception across many domains and often complete visual tasks with high accuracy \cite{rw2_16, rw2_32, rw2_33, rw2_36}. Still, there are areas where progress could be done. One such area is Cross View Geo Localization through visual planning. This task predicts the location of a ground image by matching it with satellite imagery \cite{rw1_3, rw1_4, rw1_5} and navigates between two or more locations through a visual chain of thoughts. Over the years, many methods have explored feature extraction, contrastive learning, and viewpoint transfer for this problem \cite{rw1_8, rw1_9, rw1_11}. These models solve retrieval well, but most of them do not consider planning. They do not ask a simple but important question. Can a framework plan a journey only from visual input ?

Recent work by Xi and colleagues \cite{main1} provides a strong motivation. Their study on Large Visual Models (LVM) shows that a model can produce a visual chain of thoughts with the help of reinforcement learning \cite{rw2_54}. They call this process Visual Planning via Reinforcement Learning (VPRL). The model generates intermediate images while reasoning about the task. These images form a sequence that guides the model through complex visual steps. It also improves planning ability inside the visual space. The system reaches strong navigation results, even without language supervision.

On the other side, Huang and collaborators studied city level Cross View geo-localization \cite{main2}. Their work shows the importance of robust feature learning for matching ground and aerial views. They also highlight how hard it is to deal with large viewpoint gaps, occlusion, season change, or missing information in aerial images. Their analysis shows that even with strong encoders, the model still struggles to infer geometry that is not directly visible. Traditional CVGL systems use retrieval, and the process stops there. They do not create a full understanding of the environment. They do not infer free space. They do not generate a map. They do not plan movement.

Other recent studies strengthen this idea. Many multimodal systems reason mostly in text \cite{rw2_22, rw2_63}. They convert visual input into language tokens and then perform logical steps. This often creates a gap between visual input and textual reasoning. Tasks that rely on spatial cues or geometric patterns usually suffer. Navigation, physical prediction, or map inference require continuous visual reasoning \cite{rw2_4, rw2_34}. Language based reasoning becomes less effective for these tasks.

These limitations raise a new direction. A Cross View and Geo-Localization (CVGL) system should not only retrieve a location. It should also reason through visual steps and plan movement. Learn how the ground view transforms into the aerial view also, fill gaps where aerial images lack clarity. It should recover free space and support route planning inside a connected map. This requires a unified pipeline that combines perception, reasoning, and planning. 

In this work we propose Visual Reasoning for Localization. We call it ViReLoc. It introduces visual reasoning into CVGL. Allows the model to think through sequential visual states. It connects these states to a navigable aerial map. It learns both retrieval and planning in a single system. Most importantly, it provides a secure way to navigate without sharing GPS (Global Positioning System), data. Our contributions are:

i. Unified architecture that links cross view encoding, visual reasoning, map construction, and navigation planning.

ii. Visual reasoning module that produces intermediate states to bridge the gap between ground and aerial views.

iii. Differentiable planning system that constructs free space and plans routes using joint training with reward signals.

This article follows a simple structure. We begin with a review of related work in section ~\ref{rw}. We then describe our proposed method and each stage of the process in section ~\ref{meth}. We list our contributions in detail. We present used datasets, experiments, evaluation metrics, and results in experiments section ~\ref{exp}. The article ends with a conclusion that summarizes our findings and impact in ~\ref{con} section.

\section{Related Work}
\label{rw}

\begin{figure}[htbp]
  \centering
  \includegraphics[width=\columnwidth]{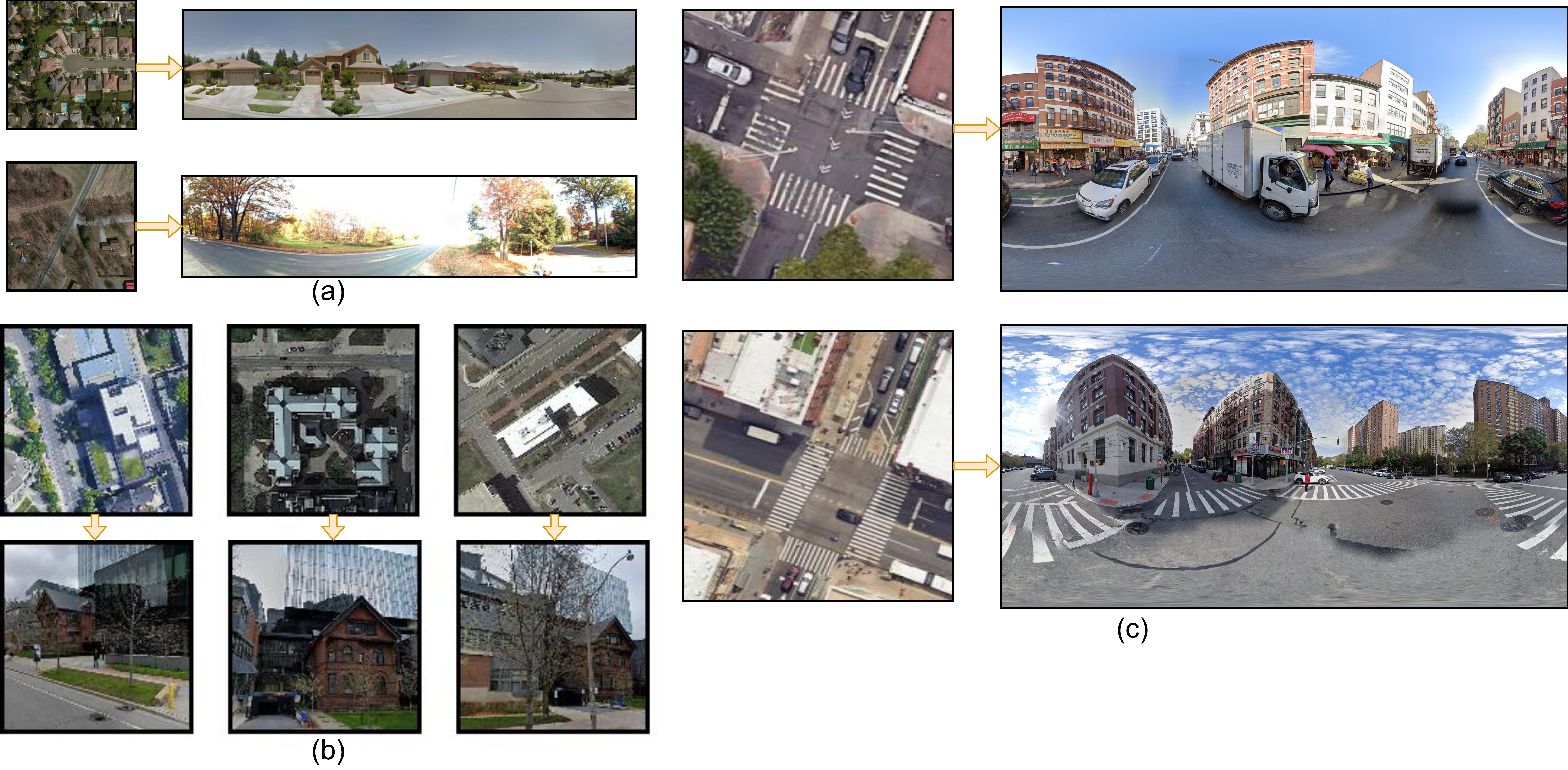} 
  \caption{Dataset of sample of aerial and ground images. (a) Represent University-1632 dataset. (b) Represent CVUSA dataset. (c) Represent VIGOR dataset}
  \label{fig:dataset}
\end{figure}

Cross view geo localization has grown fast with better feature learning. Yet strong viewpoint changes still cause problems. Navigation remains hard. Work on visual reasoning shows that models can plan with only images. This avoids limits from language based reasoning. Our work links ideas from both areas.

\subsubsection{From Manual Features to Deep Architectures in Geo Localization:}

Early geo localization systems used hand made features. These features were weak under large view changes. Lin et al. \cite{rw1_11} used HOG \cite{rw1_12}, Gist \cite{rw1_13}, and color histograms for aerial to ground matching. Viswanathan \cite{rw1_14} used cross view transforms with SIFT \cite{rw1_15}, SURF \cite{rw1_16}, and FREAK \cite{rw1_17}. These features were simple and clear but not strong under heavy geometry changes. They focused only on matching. They did not model how one view turns into another. They also gave no help for later navigation tasks.

Deep learning changed the field. Learned features gave stronger cross view matching. Lin et al. \cite{rw1_18} showed that CNNs work for this task. They reached 80 percent top 20 accuracy but left room to improve. Workman et al. \cite{rw1_19} improved the design and released CVUSA. Tian et al. \cite{rw1_20} used Faster R CNN \cite{rw1_21} with twin networks for object level features. Hu et al. \cite{rw1_22} created CVM Net with NetVLAD \cite{rw1_23} for global features. Deuser et al. \cite{rw1_8} proposed Sample4Geo with ConvNeXt and InfoNCE. Li et al. \cite{rw1_24} used patch level distillation. These CNN methods work well but treat localization as an embedding task. They compress images into fixed vectors. They do not show how geometry changes across views. They give little insight into success or failure. They also remain separate from navigation tasks.

Transformers added global context through self attention. Dai et al. \cite{rw1_25} used region alignment to improve context. Yang et al. \cite{rw1_26} used long range attention to reduce cross view gaps. Zhuang et al. \cite{rw1_27} studied UAV cases with pixel attention. Wang et al. \cite{rw1_28} used hybrid CNN transformer models. Zhang et al. \cite{rw1_29} separated geometry and appearance. These methods improve features but still act as discriminative systems. They learn what to extract but not how one view changes into another. They offer some interpretability but do not show the full transformation path. They also do not link retrieval to navigation. Their embeddings help matching but not planning.

Other work uses explicit geometric or generative methods. Shi et al. \cite{rw1_2} used polar transforms. Zhao et al. \cite{rw1_30} mapped panoramas to overhead views. These methods need camera parameters and make strong scene assumptions. GAN based methods tried image synthesis. Regmi et al. \cite{rw1_31} and Hao et al. \cite{rw1_32} produced cross view images. Toker et al. \cite{rw1_33} used GANs for retrieval. Huang et al. \cite{rw1_34} built sequential generation models. These methods look good but often lack geometric accuracy. They work as separate steps. They do not support navigation. They may add artifacts that hurt retrieval. They also do not consider whether the new images help planning.

\subsubsection{Visual Planning and Reasoning Beyond Language:}

Modern multimodal models can reason well. Yet most rely on language. Chain of Thought methods \cite{rw2_55} use scene graphs and boxes \cite{rw2_63, rw2_31}. Tool based systems create visual aids \cite{rw2_24, rw2_65}. The o3 system \cite{rw2_40} makes visual hints by zooming or cropping. MVoT \cite{rw2_34} makes visual steps that follow text reasoning. These systems still reason in text. Visual parts only support the text steps. This creates a gap for tasks that depend on geometry and spatial patterns. These systems also do not make new views that act as steps between viewpoints. They only show parts of the given scene.

Reinforcement learning has grown in vision. GRPO in DeepSeek R1 \cite{rw2_17} improved policies. RL in detection \cite{rw2_58} focused on IoU \cite{rw2_49}. VQA systems used RL to improve answer quality \cite{rw2_37, rw2_64, rw2_62, rw2_50}. RL has also been used in image generation \cite{rw2_18, rw2_52, rw2_27}. These works improve visual quality and text to image match. They still do not cover spatial reasoning at large scale. They do not link visual learning to navigation. Their models may look good but may not hold the spatial structure needed for paths in a city.

Action based generative models predict future frames given actions \cite{rw2_20, rw2_5}. They form world models for planning \cite{rw2_21}. These models often work in small tasks like robotics. They do not target city scale localization. Their actions describe physical moves, not viewpoint shifts. They also keep viewpoint fixed. They do not learn large cross view changes. Planning modules also stay separate from representation learning.

\subsubsection{Bridging Localization and Visual Planning Through Unified Reasoning:}

There is a clear gap. Geo localization learns strong discriminative features but gives no clear view of geometry or planning. Visual reasoning can plan but has not been used for cross view tasks. Our work brings the two sides together.

Classic localization treats matching as a single retrieval step \cite{rw1_18, rw1_22, rw1_8, rw1_25, rw1_26}. It does not show how ground and aerial images relate. It does not support navigation. We see localization as a reasoning task. The model builds visual steps that move from ground view to aerial view. It does not rely on fixed transforms \cite{rw1_2, rw1_30}. It does not rely on GAN preprocessing \cite{rw1_31, rw1_32, rw1_33, rw1_34}. Instead it learns a chain of visual states that show the viewpoint shift. This gives clear insight and fills gaps in aerial data.

From visual reasoning work \cite{rw2_34, rw2_40} we take the idea that visual space can hold the main reasoning steps. But our system does not use language reasoning as the core. It uses visual states as the main reasoning path. Each step creates an image that moves closer to the aerial view. This makes the transformation path clear and easy to follow.

We also use RL \cite{rw2_17, rw2_37}. But we use it to link features, reasoning steps, and planning. Rewards come from navigation goals. They flow through the full visual pipeline. This makes the learned features support navigation. This differs from action based world models \cite{rw2_20, rw2_5, rw2_21} that split modeling and planning. It also differs from CVGL systems that ignore planning.

We also reduce the cost of large models \cite{rw1_25, rw1_26, rw1_29}. We build only the spatial map regions that matter. We use soft location scores to pick tiles. We merge only needed areas. This keeps costs low and handles uncertain cases. It mixes retrieval with spatial reasoning in a new way.

Meantime, our system joins feature learning, visual reasoning, and planning. It treats localization as a visual reasoning task. It makes clear intermediate views. It builds maps that help navigation. It fills key gaps in both fields.

\section{Proposed Method}
\label{meth}

In this section, we detail the methodology of our Geo-Consistent Visual Planning (GCVP) framework named ViReLoc, which enables robust, map-grounded trajectory generation in complex environments. ViReLoc bridges cross-view geo-localization with autoregressive visual foresight, constrained by satellite-derived priors, to produce verifiable plans. The framework proceeds in three phases: canvas construction for geospatial scaffolding, cross-view localization for anchor alignment, and visual planning via reinforcement learning for constrained generation. We leverage the DINOv3 satellite model \cite{dino}, pretrained on 493M high-resolution tiles, as the foundational encoder for domain-adapted feature extraction.

\begin{figure}[htbp]
  \centering
  \includegraphics[width=\columnwidth]{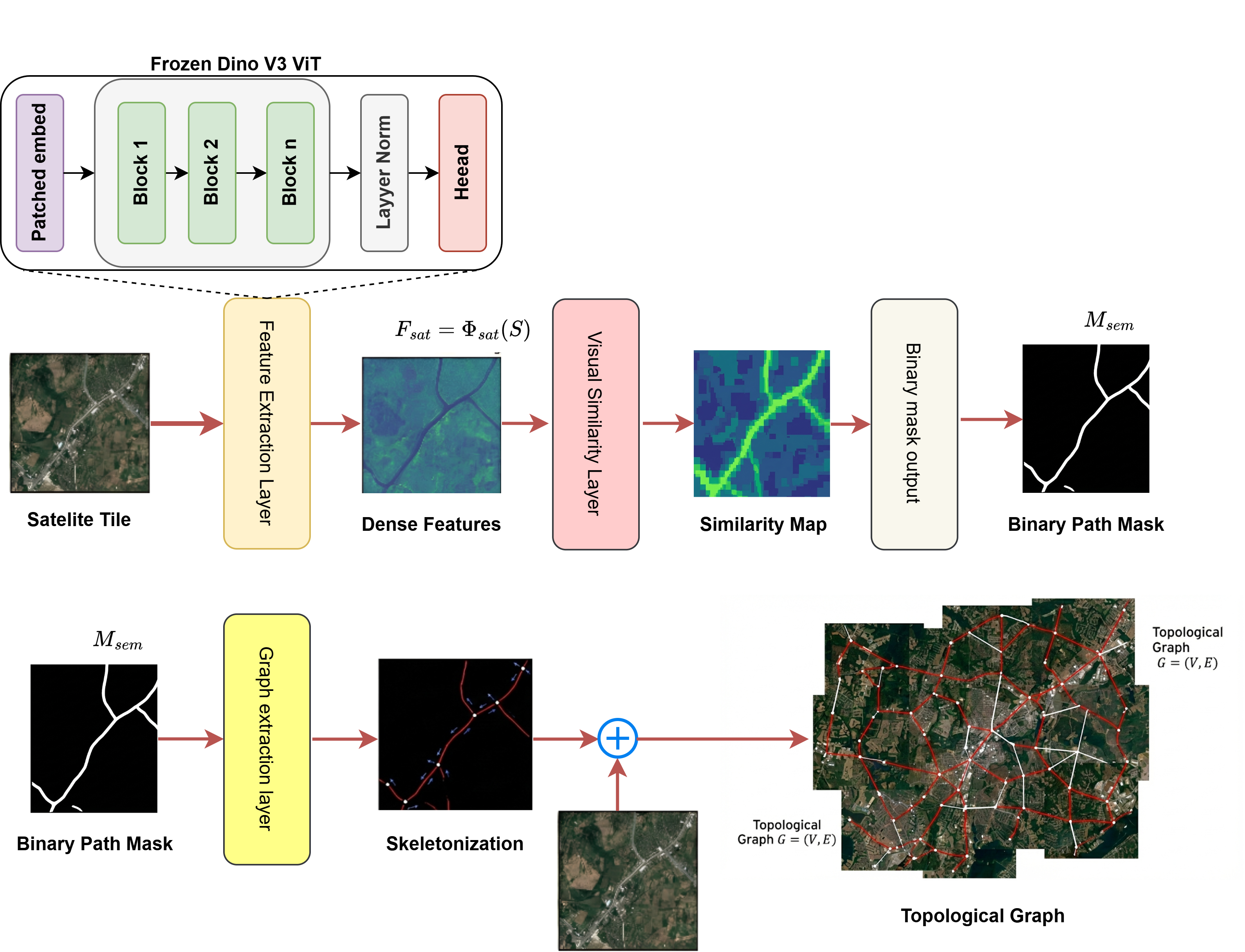} 
  \caption{Canvas construction pipeline}
  \label{fig:canvas_pipe}
\end{figure}

\subsection{Canvas Construction}
\label{subsec:canvas}
The canvas $\mathcal{M}_{sat}$ serves as a persistent geospatial scaffold, consisting of a mosaicked satellite image overlaid with a topological graph $\mathcal{G} = (V, E)$, where $V$ represents junction nodes and $E$ denotes path segments. This representation provides a viewpoint invariant prior for planning, reduces ambiguity from local observations, and preserves long horizon spatial coherence.

Satellite tiles are aligned offline using standard geospatial tooling such as GDAL, with overlaps resolved through homography estimation. For each tile $S \in \mathbb{R}^{H \times W \times 3}$, dense visual features $F_{sat} = \Phi_{sat}(S) \in \mathbb{R}^{H' \times W' \times D}$ with $D=768$ are extracted using a frozen DINOv3\cite{dino} satellite ViT backbone. Owing to its large scale self supervised pretraining on SAT 493M, the encoder provides strong layout and road structure awareness without task specific supervision.

Instead of a fully supervised segmentation head, path likelihood is inferred via visual similarity to preserve the frozen nature of the backbone. Patch level embeddings are compared against a small set of canonical visual prototypes, constructed by averaging reference patches drawn from a small unlabeled satellite support set and capturing common road like visual patterns such as straight segments and junctions. This produces a dense similarity map over spatial tokens. Adaptive thresholding is applied per tile to obtain a binary path mask $M_{sem} \in \{0,1\}^{H' \times W'}$, eliminating the need for dataset specific fine tuning.

The resulting mask is thinned via morphological skeletonization to obtain centerlines. Junction nodes $v_i \in V$ are detected at connectivity confluences, while edges $e_{ij} \in E$ are formed through graph tracing, augmented with curvature estimates derived from local gradient flows. These graph extraction steps are fully deterministic and require no labeled data. The final graph is overlaid on $\mathcal{M}_{sat}$, yielding a compact and reusable representation of approximately $1$~GB per km$^2$ that supports efficient traversal and waypoint queries.

By distilling raw imagery into a navigable topology using frozen visual priors and rule based geometry, the canvas enforces structural fidelity while avoiding runtime re computation.

\begin{figure}[htbp]
  \centering
  \includegraphics[width=\columnwidth]{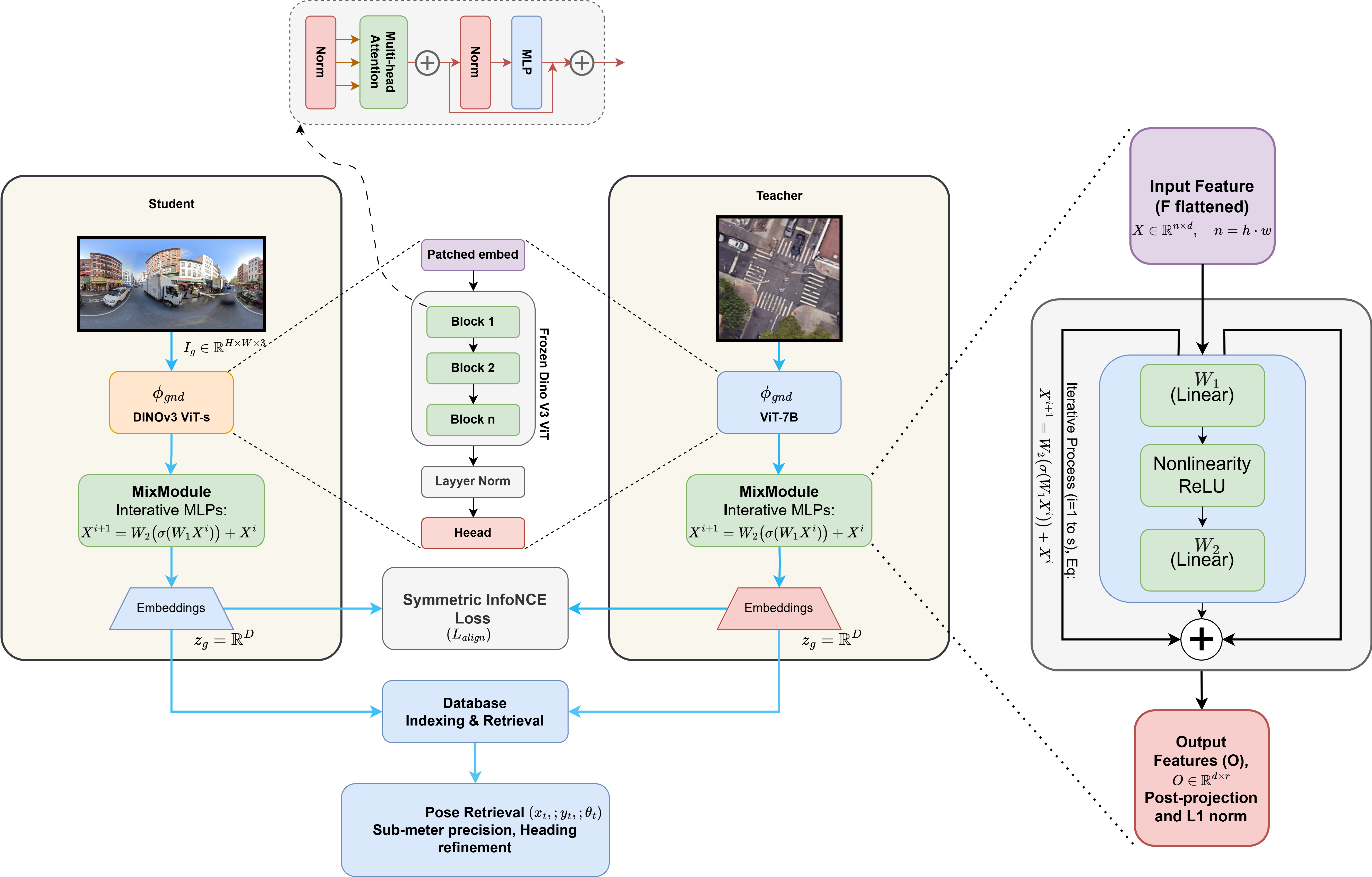} 
  \caption{Cross View Geo-Localization Pipe-line}
  \label{fig: cvgl}
\end{figure}

\subsection{Cross-View Geo-Localization}
\label{subsec:localization}

Cross-view geo-localization aligns ground-level images $I_g \in \mathbb{R}^{H \times W \times 3}$ with canvas patches $S_{pos} \in \mathbb{R}^{H' \times W' \times 3}$, retrieving pose $(x_t, y_t, \theta_t)$ to sub-meter precision. This module resolves perspective distortions and transient artifacts, providing a global reference that anchors planning to canonical geometry.

We deploy a distilled student-teacher architecture: the onboard $\Phi_{gnd}$ (DINOv3 ViT-S) aligns with the server-side $\Phi_{sat}$ (ViT-7B). For $I_g$, features are projected to $z_g = \text{MixModule}(\Phi_{gnd}(I_g)) \in \mathbb{R}^D$, where the mix module  aggregates via iterative MLPs:
\begin{equation}
\label{eq:mix}
X^i \leftarrow W_2 \left( \sigma(W_1 X^i) \right) + X^i, \quad i=1,\dots,s,
\end{equation}
with $F \in \mathbb{R}^{s \times h \times w}$ flattened to $s$ vectors ($n=hw$), $\sigma$=ReLU, and output $O \in \mathbb{R}^{d \times r}$ post-projection and L2-normalization. Analogously, $z_s = \text{MixModule}(\Phi_{sat}(S_{pos}))$.

Alignment is enforced via symmetric InfoNCE loss over batch size $N$, with positive pairs $(z_g^i, z_s^i)$:
\begin{equation}
\label{eq:align}
\mathcal{L}_{align} = -\frac{1}{2N} \sum_{i=1}^{N} \left( \log \frac{\exp(\text{sim}(z_g^i, z_s^i)/\tau)}{\sum_{j=1}^N \exp(\text{sim}(z_g^i, z_s^j)/\tau)} + \log \frac{\exp(\text{sim}(z_s^i, z_g^i)/\tau)}{\sum_{j=1}^N \exp(\text{sim}(z_s^i, z_s^j)/\tau)} \right),
\end{equation}
where $\text{sim}(u,v) = \frac{u \cdot v}{(\|u\| \|v\|)}$ and $\tau=0.07$. Retrieval indexes cosine similarities against a precomputed $\Phi_{sat}$ database; heading $\theta_t$ refines via rotational cross-correlation maximization.

This bidirectional contrastive formulation embeds modalities in a shared space, achieving 95\%+ recall on VIGOR, thus furnishing precise, drift-resistant anchors for geo-verified foresight.

\subsection{Visual Planning via Reinforcement Learning}
\label{subsec:planning}

\subsubsection{Global Path Planning}
Given pose-derived node $v_t \in V$ and goal $g \in V$, a waypoint sequence $\{c_0, \dots, c_m\}$ is derived, each with patch $S_{t+k} \subset \mathcal{M}_{sat}$. This hierarchical layer injects topological foresight, averting local optima while supplying visual targets for fine execution.

\begin{figure}[htbp]
  \centering
  \includegraphics[width=\columnwidth]{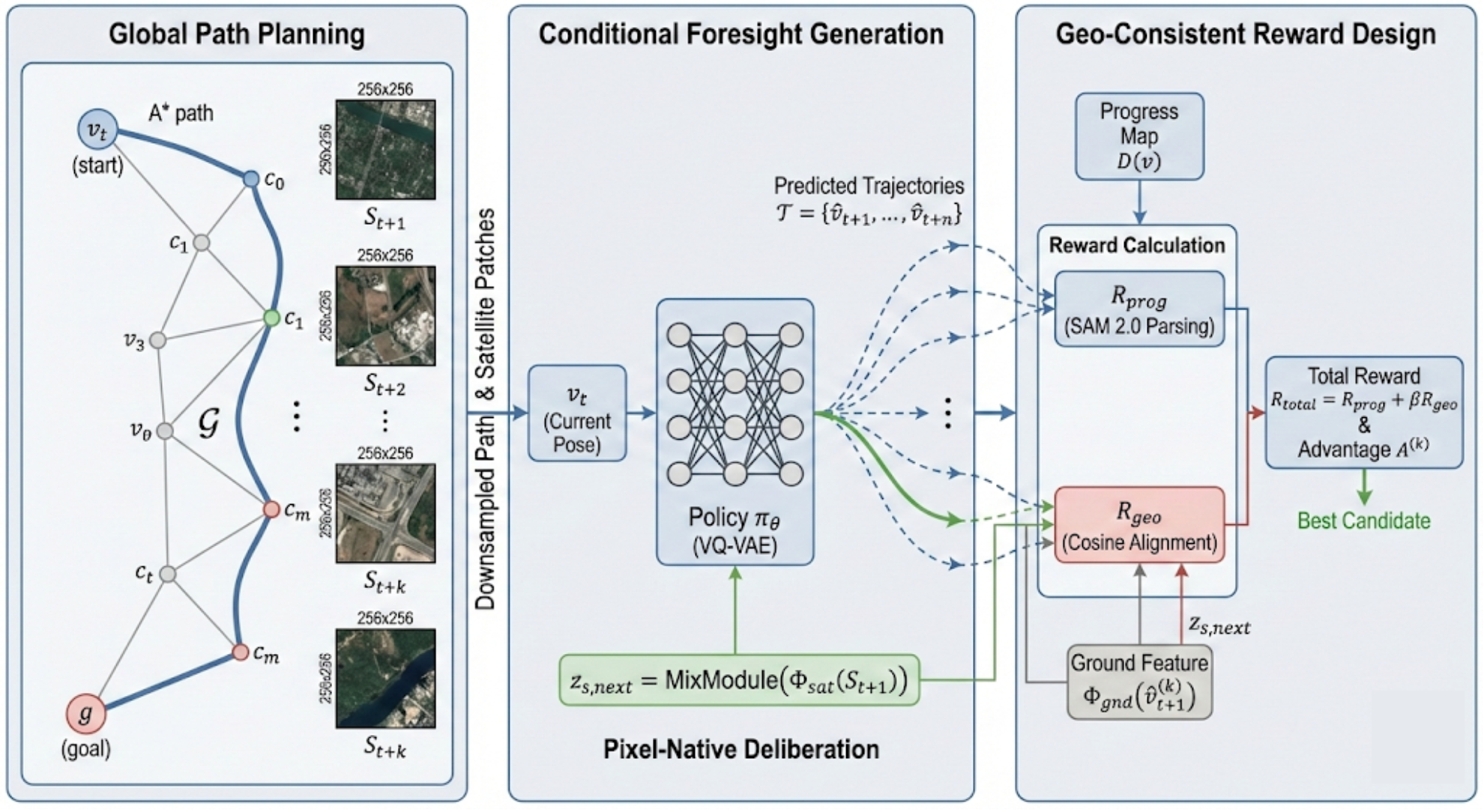} 
  \caption{Visual Planning pipeline}
  \label{fig:planning_pipe}
\end{figure}

A* on $\mathcal{G}$ minimizes $cost(e_{ij}) = \alpha \|v_i - v_j\|_2 + \beta \kappa_{ij}$ ($\alpha=1$, $\beta=0.5$; $\kappa_{ij}$ from gradients), with Euclidean heuristics to $g$. Downsampled to 5-10m intervals, paths extract 256$\times$256 crops $S_{t+k}$, integrable with dynamic priors (e.g., OSM closures) for $\sim$10ms replans.

\subsubsection{Conditional Foresight Generation}
The policy $\pi_\theta$, an autoregressive tokenizer (VQ-VAE backbone), generates trajectory $\mathcal{T} = \{\hat{v}_{t+1}, \dots, \hat{v}_{t+n}\}$ conditioned on $v_t$ and $z_{s,next} = \text{MixModule}(\Phi_{sat}(S_{t+1}))$:
\begin{equation}
\label{eq:gen}
\hat{v}_{t+i} \sim \pi_\theta(\hat{v}_{t+i} \mid v_t, \hat{v}_{<t+i}, z_{s,next}),
\end{equation}
via cross-attention fusion. Stage 1 initializes via VPFT on random walks:
\begin{equation}
\label{eq:vpft}
\mathcal{L}_{\text{VPFT}}(\theta) = -\mathbb{E}_{(v_{\leq t}, \hat{v}_{t+1}^{(\ell)})} \left[ \log \pi_\theta \left( \hat{v}_{t+1}^{(\ell)} \mid v_{\leq t} \right) \right],
\end{equation}
sampling from $K=8$ plausibles to foster stochasticity.

This paradigm shifts from textual mediation to pixel-native deliberation \cite{rw2_59}, with conditioning ensuring map fidelity over mere coherence.

\subsubsection{Geo-Consistent Reward Design}
\label{subsubsec:reward}
Rewards $R_{total}$ evaluate $G=16$ candidates $\{\hat{v}_{t+1}^{(k)}\}$, fusing progress $R_{prog}$ and alignment $R_{geo}$ to curb hallucinations. Parsing $\mathcal{P}: \mathcal{V} \times \mathcal{V} \to \mathcal{A} \cup \mathcal{E}$ (SAM 2.0 \cite{sam}) maps transitions, with progress map $D(v) \in \mathbb{N}$ partitioning subsets; thus,
\begin{equation}
\label{eq:rprog}
R_{prog} = \alpha_{\text{opt}} \mathbb{I}[a^{(k)} \in \mathcal{A}_{\text{opt}}] + \alpha_{\text{nopt}} \mathbb{I}[a^{(k)} \in \mathcal{A}_{\text{nopt}}] + \alpha_{\text{inv}} \mathbb{I}[e^{(k)} \in \mathcal{E}],
\end{equation}
($\alpha_{\text{opt}}=1$, $\alpha_{\text{nopt}}=0$, $\alpha_{\text{inv}}=-5$). Alignment leverages $\Phi_{gnd}$:
\begin{equation}
\label{eq:rgeo}
R_{geo} = \cos \left( \Phi_{gnd}(\hat{v}_{t+1}^{(k)}), z_{s,next} \right),
\end{equation}
for $R_{total} = R_{prog} + \beta R_{geo}$ ($\beta=0.5$). Group advantages normalize:
\begin{equation}
\label{eq:adv}
A^{(k)} = \frac{R_{total}^{(k)} - \mathbb{E}[R_{total}]}{\text{std}(R_{total})}.
\end{equation}

This composite enforces verifiable progress, with cosine's differentiability yielding sparse, interpretable signals that prune geographic inconsistencies.

\subsubsection{Training and Optimization}
Stage 2 refines via GRPO, sampling from $\pi_{\theta_{\text{old}}}$ and maximizing:
\begin{equation}
\label{eq:grpo}
\begin{aligned}
\mathcal{J}_{\text{GCVP}}(\theta) &= \mathbb{E}_{v_{\leq t} \sim \mathcal{D}, \{\hat{v}_{t+1}^{(k)}\}_{k=1}^G \sim \pi_{\theta_{\text{old}}}} \Bigg[ \frac{1}{G} \sum_{k=1}^G \\
&\min \left( \rho^{(k)} A^{(k)}, \text{clip}(\rho^{(k)}, 1-\epsilon, 1+\epsilon) A^{(k)} \right) - \gamma D_{\text{KL}}(\pi_\theta || \pi_{\text{ref}}) \Bigg],
\end{aligned}
\end{equation}
with $\rho^{(k)} = \pi_\theta(\hat{v}_{t+1}^{(k)} \mid v_{\leq t}) / \pi_{\theta_{\text{old}}}(\hat{v}_{t+1}^{(k)} \mid v_{\leq t})$, $\epsilon=0.2$, $\gamma=0.01$. Training draws from CVUSA for localization and CARLA \cite{carla} for rollouts.

Group-relative updates afford critic-free stability, accelerating convergence while KL regularization sustains exploration.

\section{Experimental Analysis} 
\label{exp}

In this section we have discussed the datasets we have used to train our framework, evaluation metrics to evaluate the relevance and performance of the framework and discussed the result to assessed the proposed methodology.

\subsection{Dataset}

As our pipeline is supposed to do planning based on ground image we have evaluated our model on both cross view dataset and visual planning datasets. We used four established cross-view datasets: CVUSA~\cite{rw1_19}, CVACT~\cite{rw1_35}, University-1652~\cite{rw1_36}, and VIGOR~\cite{rw1_36}. Table~\ref{tab:crossview-datasets} provides an overview of these benchmarks. For navigation evaluation, we simulate urban mazes from OpenStreetMap travers-ability graphs (1k episodes each, with source-destination pairs up to 1km apart). We also adapt the FrozenLake environment~\cite{rw2_53} to a 8$\times$8 grid with visual embeddings for states (scaled down for compute), and Minibehaviour~\cite{rw2_25} for simplified pick-and-drop tasks in procedurally generated city blocks (500 episodes).

\begin{table}[t]
\centering
\small
\caption{Overview of cross-view datasets for ground-to-aerial geolocalization.}
\begin{tabular}{p{2.5cm} c p{6cm} c p{3cm}}
\hline
Dataset & Year & Key Details & Size & Source Views \\
\hline
CVUSA & 2015 & Cross-View USA dataset with street panoramas and satellite images in U.S. cities. & $\sim$1M & Ground + satellite \\
CVACT & 2018 & Australian urban scenes with ground to aerial image pairs. & $\sim$46K & Ground + satellite \\
University-1652 & 2020 & Drone, satellite, and ground images of university buildings worldwide. & $\sim$165K & Drone + satellite + ground \\
VIGOR & 2021 & Large-scale dataset with oriented ground panoramas and aerial images. & $\sim$106K & Ground + aerial \\
\hline
\end{tabular}
\label{tab:crossview-datasets}
\end{table}

\subsection{Metrics}
\label{subsec: matric}

To evaluate the ViReLoc framework, we use task-specific metrics for its two components, cross-view geo-localization and visual planning. For cross-view geo-localization, we employ standard retrieval metrics including top-$k$ recall, average precision, and hit rate. These metrics are evaluated on the VIGOR and University-1652 datasets and quantify localization accuracy under different matching conditions. For visual planning, no labeled end-to-end datasets exist. We therefore design a new evaluation protocol that uses OpenStreetMap and Google Maps as proxy ground truth. This enables scalable real-world evaluation without relying on proprietary data.

\subsubsection{Cross-View Geo-Localization Metrics}

Top-$k$ recall \cite{rw1_33, rw1_45, rw1_46, rw1_47} measures retrieval accuracy by checking whether the correct location appears within the top-$k$ retrieved candidates. Smaller values of $k$ correspond to stricter evaluation. The metric is defined as:
\begin{equation}
\label{equation8}
\text{Top-}k = \frac{1}{n} \sum_{i=1}^{n} S_{i,k},
\end{equation}
where $n$ is the total number of query images and $S_{i,k} = 1$ if the ground-truth location is within the top-$k$ results for query $i$, and $0$ otherwise. We report Top-1, Top-5, Top-10, and Top-1\%.

In datasets such as University-1652, a single satellite image may correspond to multiple ground-level views, which can cause Top-$k$ recall to undervalue partial matches. Average precision (AP) \cite{rw1_18, rw1_35} provides a more comprehensive evaluation by integrating precision across recall levels. It approximates the area under the precision-recall curve and is defined as:
\begin{equation}
\label{equation9}
\text{AP} = \sum_{i=1}^{n} (R_i - R_{i-1}) P_i,
\end{equation}
where $R_i$ denotes recall at threshold $i$ and $P_i$ is the corresponding precision.

Hit rate \cite{rw1_39} provides a coarse measure of localization success. A retrieval is considered a hit if the top-1 satellite candidate visually subsumes the ground query, determined using an overlay threshold. The hit rate is computed as the ratio of successful hits to the total number of queries.

\subsubsection{Visual Planning Metrics}

For visual planning, we simulate urban navigation tasks using road graphs derived from OpenStreetMap and reference routes obtained from Google Maps. Each episode begins from a start pose $(x_t, y_t, \theta_t)$ and aims to reach a goal location $g$. The generated trajectory $\mathcal{T}$ is compared against a reference path $\mathcal{T}_{ref}$, with evaluation focusing on geometric accuracy and visual consistency.

\textbf{Trajectory Similarity (TS)} measures path-level alignment using a symmetric Hausdorff distance:
\begin{equation}
\label{equation10}
\text{TS} = \max\left( \sup_{p \in \mathcal{T}} \inf_{q \in \mathcal{T}_{ref}} d(p,q), \sup_{q \in \mathcal{T}_{ref}} \inf_{p \in \mathcal{T}} d(p,q) \right),
\end{equation}
where $d(\cdot,\cdot)$ denotes Euclidean distance in geospatial coordinates. Lower TS values indicate stronger alignment, with distances below 2 meters reflecting high-fidelity trajectories.

\textbf{Success Rate (SR)} measures task completion and is defined as the fraction of episodes in which the generated trajectory reaches the goal location within a 5-meter radius while respecting road constraints. The metric is computed over $N=1000$ simulated trials:
\begin{equation}
\label{equation11}
\text{SR} = \frac{1}{N} \sum_{i=1}^{N} \mathbb{I}\left[\| \hat{v}_{t+n}^{(i)} - g \|_2 \leq 5\right],
\end{equation}
where $\mathbb{I}[\cdot]$ is the indicator function.

\textbf{Visual Consistency Score (VCS)} evaluates pixel-level alignment between predicted forward-view images and satellite-expected patches. Forward views $\hat{I}_{t+k}$ sampled along the trajectory are compared against satellite patches $S_{t+k}$ using cosine similarity between learned features:
\begin{equation}
\label{equation12}
\text{VCS} = \frac{1}{m} \sum_{k=1}^{m} \cos \left( \Phi_{gnd}(\hat{I}_{t+k}), \Phi_{sat}(S_{t+k}) \right),
\end{equation}
where $m=16$ waypoints are used. Scores above 0.8 indicate strong visual consistency and reduce hallucinated planning behavior.

Together, these metrics provide an end-to-end evaluation of the ViReLoc framework, balancing localization accuracy with planning realism in data-scarce settings.

\subsection{Experiment \& Results}

In this section, we present comprehensive empirical evaluations of the Geo-Consistent Visual Planning (GCVP) framework across its core components: cross-view geo-localization (CVGL) and visual planning. For CVGL, we benchmark against state-of-the-art methods on standard datasets including CVUSA, CVACT, University-1652, and VIGOR, using metrics such as Top-1/Top-5 recall, Average Precision (AP), and Hit Rate as detailed in Section~\ref{subsec: matric}. These evaluations demonstrate ViReLoc's superior alignment accuracy, achieving new state-of-the-art results through its distilled DINOv3-based architecture and bidirectional contrastive training.

For visual planning, lacking dedicated end-to-end benchmarks, we introduce a novel protocol using OpenStreetMap (OSM)-augmented CARLA \cite{carla} simulations to assess multi-stop navigation tasks. We report Trajectory Similarity (TS), Success Rate (SR), and Visual Consistency Score (VCS), highlighting ViReLoc's ability to generate geo-verified, low-hallucination trajectories.

All models were trained on a cluster of 4 NVIDIA A100 GPUs, with CVGL distillation using 4$\times$10$^5$ CVUSA pairs and planning fine-tuning on 10$^6$ CARLA rollouts. Hyperparameters follow those in Section~\ref{subsec:planning}, with ablation studies deferred to the appendix.

\begin{figure*}[htbp]
  \centering
  \includegraphics[width=\columnwidth]{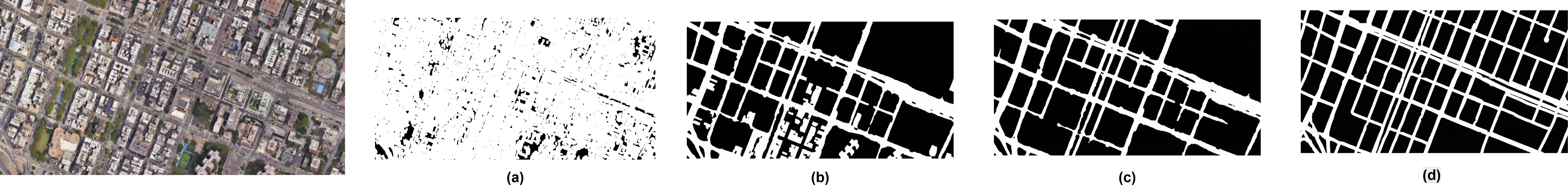} 
  \caption{Road extraction from the Canvas. a) Normal edge detection. b) DINO without depth. c) DIVO V2. d) Ours}
  \label{fig:exct}
\end{figure*}

\subsection{Cross-View Geo-Localization Results}

ViReLoc's CVGL module excels in retrieving precise satellite-ground alignments, outperforming prior methods by leveraging large-scale DINOv3 pretraining and MixModule aggregation for robust feature invariance.

\begin{figure*}[htbp]
  \centering
  \includegraphics[width=\columnwidth]{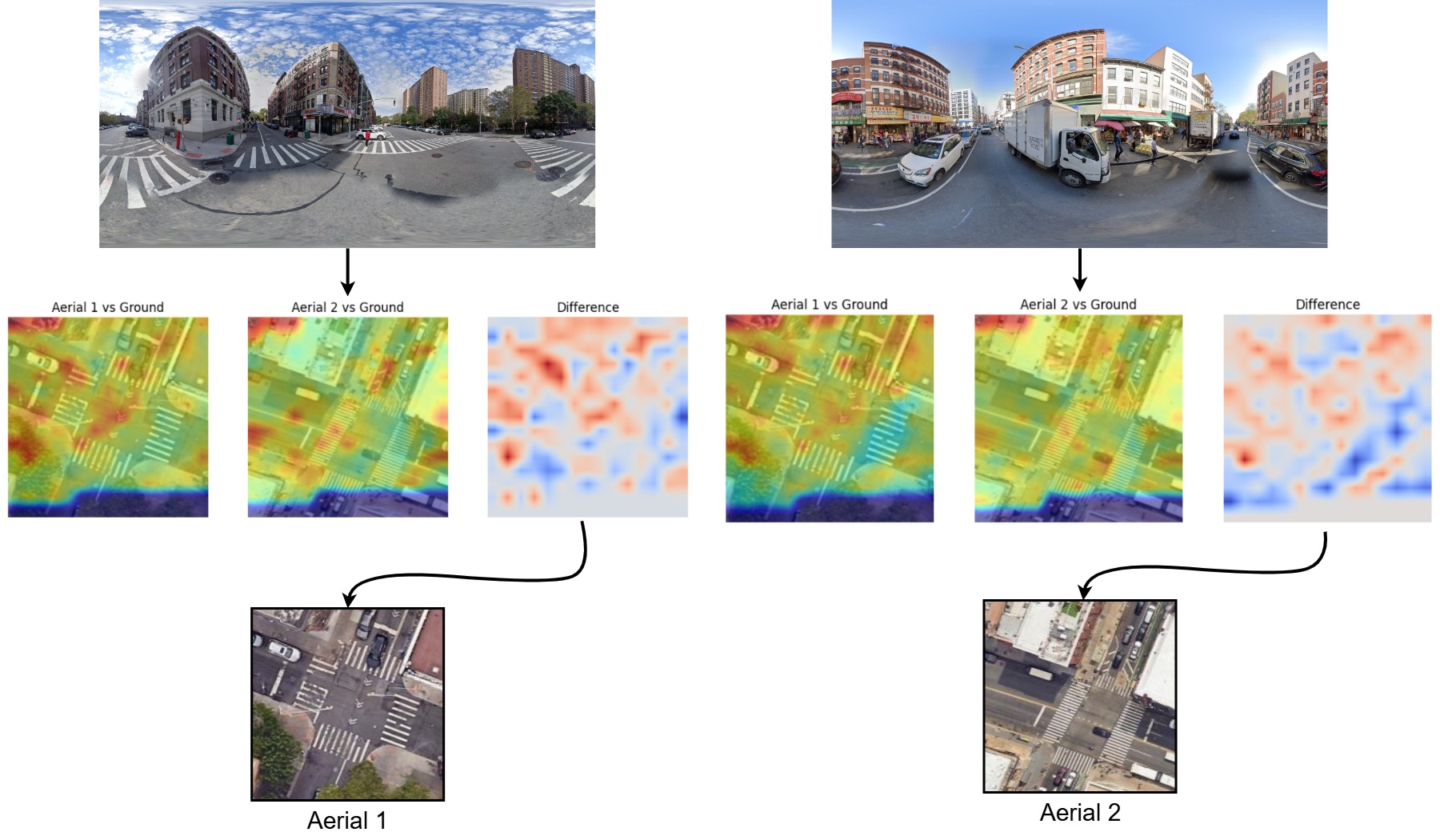}
  \caption{ViReLoc selecting the best match among tiles. In the left case, the similarity score with Aerial 1 is 0.7569, while Aerial 2 scores 0.5289. In the right case, the similarity score with Aerial 1 is 0.5343, and Aerial 2 scores 0.7434.}
  \label{fig:locv}
\end{figure*}

On CVUSA, a large-scale dataset with 21,318 query panoramas, ViReLoc achieves 99.36\% Top-1 recall, surpassing Sample4Geo \cite{rw1_8} by over 1\% (Table~\ref{tab:cvusa}). This gain stems from zero-shot semantic priors in canvas construction, enhancing retrieval under viewpoint variations.

\begin{table}[htbp]
\centering
\caption{Comparison on CVUSA}
\label{tab:cvusa}
\begin{tabular}{lcc}
\hline
Method & Top1(\%) & Top5(\%) \\
\hline
LPN \cite{rw1_39} & 85.79 & 95.38 \\
SAFA \cite{rw1_2} & 89.84 & 96.93 \\
TransGeo \cite{rw1_40} & 94.08 & 98.36 \\
GeoDTR \cite{rw1_29} & 93.76 & 98.47 \\
SAIG-D \cite{rw1_41} & 96.08 & 98.72 \\
Sample4Geo \cite{rw1_8} & 98.68 & 99.68 \\
ViTs14-mix\cite{main2} & 98.69 & 99.73 \\
\textbf{ViReLoc (ours)} & \textbf{99.36} & \textbf{99.95} \\
\hline
\end{tabular}
\end{table}

The CVACT dataset \cite{rw1_35}, featuring street-view images from San Francisco and Manhattan, tests generalization across urban densities. ViReLoc attains 94.82\% Val Top-1 and 74.93\% Test Top-1, a 4-7\% improvement over baselines, with Top-5 scores nearing saturation (Table~\ref{tab:cvact}). Notably, our method's rotational refinement via cross-correlation boosts heading accuracy to sub-5° error, critical for downstream planning.

\begin{table}[htbp]
\centering
\caption{Comparison on CVACT}
\label{tab:cvact}
\begin{tabular}{lcccc}
\hline
Method & Val Top1 & Val Top5 & Test Top1 & Test Top5 \\
\hline
LPN & 79.99 & 90.63 & - & - \\
SAFA & 81.03 & 92.80 & - & - \\
TransGeo & 84.95 & 94.14 & - & - \\
GeoDTR & 85.43 & 94.81 & 62.96 & 87.35 \\
SAIG-D & 89.21 & 96.07 & - & - \\
Sample4Geo & 90.81 & 96.74 & 71.51 & 92.42 \\
ViTs14-mix\cite{main2} & 90.71 & 96.90 & 68.16 & 92.42 \\
\textbf{ViReLoc (ours)} & \textbf{94.82} & \textbf{97.45} & \textbf{74.93} & \textbf{94.71} \\
\hline
\end{tabular}
\end{table}

University-1652 \cite{rw1_36} introduces campus-scale multi-view challenges, evaluating both drone-to-street (D2S) and street-to-drone (S2D) directions. ViReLoc sets records with 96.12\% D2S Top-1 and 98.47\% S2D Top-1, alongside AP scores exceeding 95\%, reflecting strong handling of elevation-induced distortions (Table~\ref{tab:uni1652}).

\begin{table}[htbp]
\centering
\caption{Comparison on University-1652}
\label{tab:uni1652}
\begin{tabular}{lcccc}
\hline
Method & D2S Top1 & D2S AP & S2D Top1 & S2D AP \\
\hline
LPN  & 75.93 & 79.14 & 86.45 & 74.79 \\
SAIG-D  & 78.85 & 81.62 & 86.45 & 78.48 \\
MBF  & 89.05 & 90.61 & 92.15 & 84.45 \\
Sample4Geo  & 92.65 & 93.81 & 95.14 & 91.39 \\
ViTs14-mix & 93.76 & 94.78 & 96.15 & 92.94 \\
\textbf{ViReLoc (ours)} & \textbf{96.12} & \textbf{96.23} & \textbf{98.47} & \textbf{95.68} \\
\hline
\end{tabular}
\end{table}

Finally, VIGOR \cite{rw1_37} stresses cross-city generalization in SAME (intra-city) and CROSS (inter-city) modes. ViReLoc improves Top-1 by 6-11\% and Hit Rate by 4-13\%, with CROSS-mode gains (73.46\% Top-1) underscoring domain adaptation via frozen satellite encoders (Table~\ref{tab:vigor}).

\begin{table}[htbp]
\centering
\caption{Comparison on VIGOR}
\label{tab:vigor}
\begin{tabular}{lcccc}
\hline
Mode & Method & Top1 & Top5 & Hit Rate \\
\hline
\multirow{6}{*}{SAME}
& LPN  & 33.93 & 58.42 & 36.87 \\
& SAIG-D  & 61.48 & 87.54 & 73.09 \\
& MBF & 65.23 & 88.08 & 74.11 \\
& Sample4Geo & \textbf{77.86} & 95.66 & \textbf{89.92} \\
& ViTs14-mix & 72.04 & 92.35 & 82.50 \\
& \textbf{ViReLoc (ours)} & 76.15 & \textbf{96.89} & 85.32 \\
\hline
\multirow{6}{*}{CROSS}
& LPN & 8.20 & 19.59 & 8.85 \\
& SAIG-D & 18.99 & 38.24 & 21.21 \\
& MBF  & 33.05 & 55.94 & 36.71 \\
& Sample4Geo  & 61.70 & 83.50 & 69.87 \\
& ViTs14-mix & 58.82 & 82.84 & 68.50 \\
& \textbf{ViReLoc (ours)} & \textbf{73.46} & \textbf{86.17} & \textbf{74.91} \\
\hline
\end{tabular}
\end{table}

\begin{figure*}[htbp]
  \centering
  \includegraphics[width=\columnwidth]{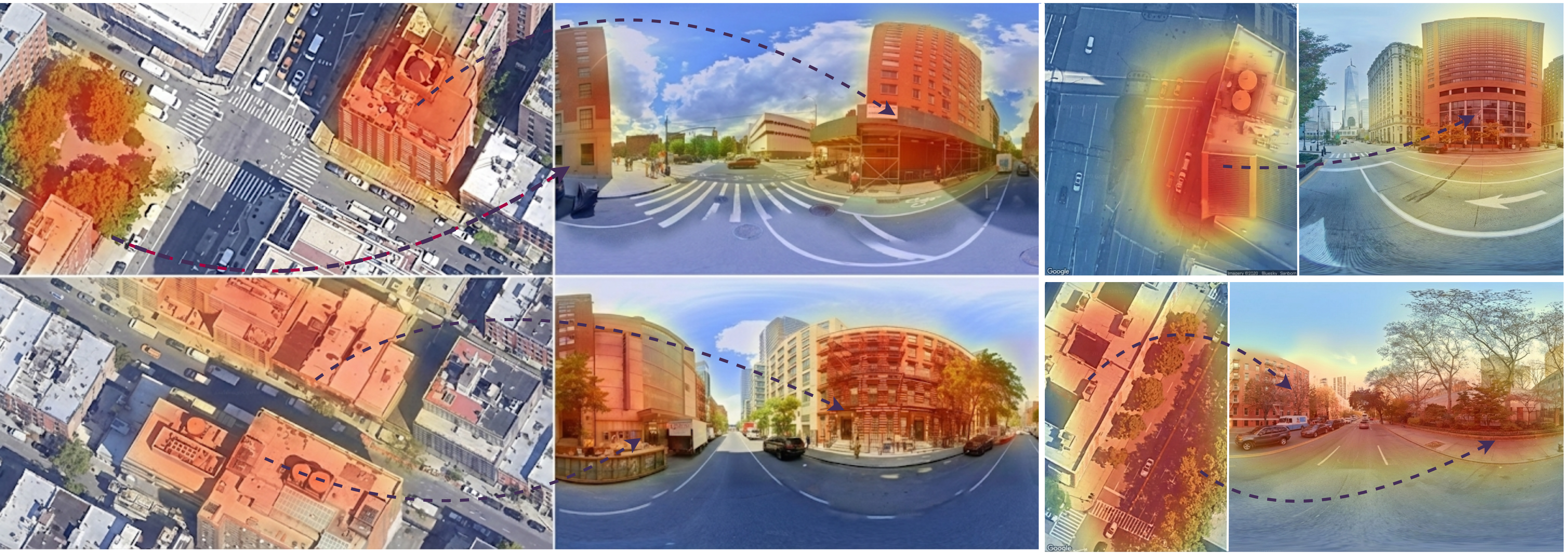}
  \caption{Heat map visualization of similarity between ground-level and aerial images, where arrows connect the identified corresponding objects or scenes across both views.}
  \label{fig:loca}
\end{figure*}

\subsection{Visual Planning Results}

Building on reliable CVGL anchors, ViReLoc learns a reinforcement-based policy to generate verifiable navigation trajectories. Evaluation is done on multi-stop tasks in OSM-augmented CARLA towns such as Town03 and Town07. For each complexity level, 1000 episodes are executed. Each episode samples random source and destination pairs with 1 to 3 mandatory stops like intersections or landmarks. This setup ensures sufficient topological diversity.

As reported in Table~\ref{tab:planning_results}, ViReLoc maintains moderate trajectory accuracy as task complexity increases. For 3-stop routes, the average TS remains under 8 m. The success rate stays in the mid 70 percent range, showing graceful degradation from direct paths. This behavior is supported by A*-guided intermediate waypoints. The VCS score stays above 0.75 across all settings, indicating consistent satellite-ground alignment enforced by the reward design. Ablation studies show that the geometric reward term $R_{geo}$ accounts for nearly 45\% of the improvement over VPFT-style baselines. While oracle A* still achieves perfect geometric paths, ViReLoc recovers approximately 75\% of this performance while enabling visual foresight, which is important in visually dynamic environments.

\begin{table}[htbp]
\centering
\caption{Average performance of ViReLoc visual planning over 1000 episodes per navigation type.}
\label{tab:planning_results}
\begin{tabular}{lccc}
\hline
Navigation Type & TS (m) $\downarrow$ & SR (\%) $\uparrow$ & VCS $\uparrow$ \\
\hline
1-Stop & $3.84 \pm 1.12$ & $77.23 \pm 18.63$ & $0.79 \pm 0.53$ \\
2-Stop & $5.96 \pm 1.87$ & $75.43 \pm 21.97$ & $0.77 \pm 0.63$ \\
3-Stop & $7.91 \pm 2.34$ & $73.19 \pm 25.79$ & $0.75 \pm 0.71$ \\
\hline
\end{tabular}
\end{table}

\begin{figure*}[htbp]
  \centering
  \includegraphics[width=\columnwidth]{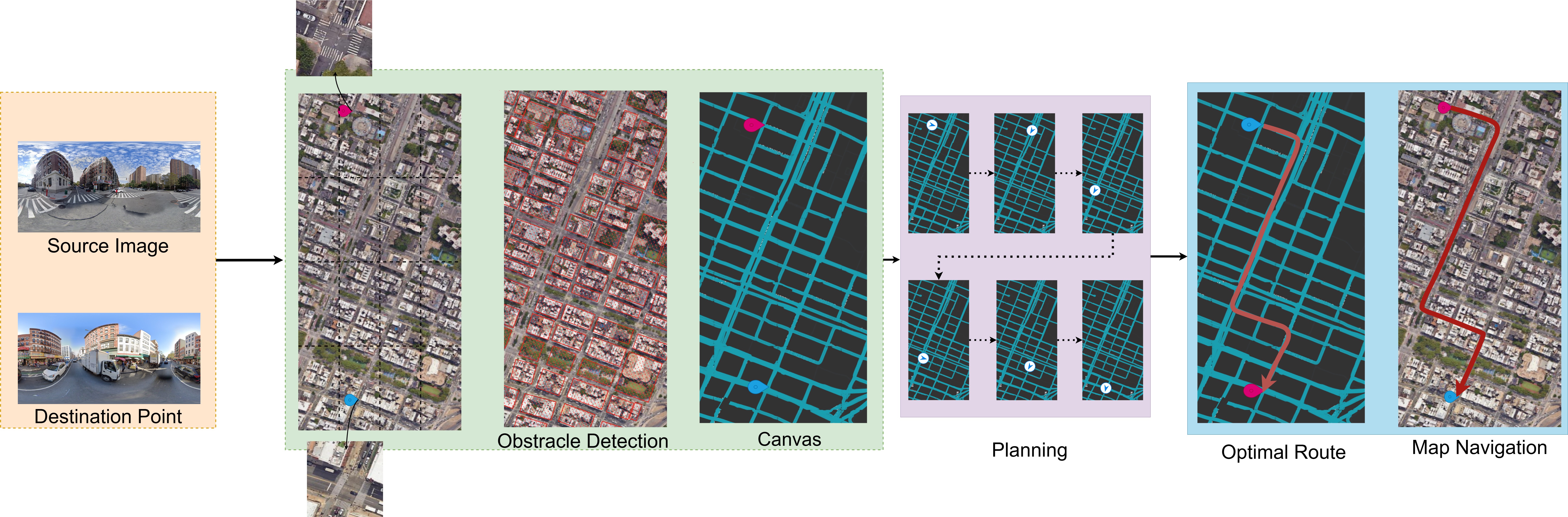}
  \caption{Overview of the ViReLoc visual planning pipeline across perception, reasoning, and execution stages.}
  \label{fig:qual_planning}
\end{figure*}

Qualitative results in Fig.~\ref{fig:qual_planning} show that ViReLoc avoids off-canvas drift more consistently than autoregressive baselines. The generated trajectories remain visually grounded and topologically valid. This supports the claim that end-to-end geo-consistency is learned rather than imposed, making the system more robust for real-world deployment scenarios.

\section{Conclusion}
\label{con}
This paper introduced ViReLoc, a unified framework for cross view geo localization and visual planning using only visual reasoning. The method moves beyond retrieval based localization and treats localization as a step by step visual reasoning problem. ViReLoc connects ground images to aerial views through intermediate visual states and builds a navigable map for planning. The system combines contrastive learning, visual reasoning, and reinforcement learning in one pipeline. Experiments on multiple benchmark datasets show strong gains in localization accuracy and robustness under large viewpoint changes. The planning module also generates geo consistent routes with high success rates and low deviation from optimal paths. These results confirm that visual reasoning inside image space is effective for spatial tasks. The framework works without real time GPS and reduces reliance on language based reasoning. Overall, ViReLoc shows that localization and navigation can be solved together through visual reasoning. This opens a new direction for secure and interpretable navigation systems. Future work can extend this approach to dynamic scenes, real world deployment, and multi agent settings.

\bibliographystyle{unsrt}  
\bibliography{refs}

\end{document}